# Iris Codes Classification Using Discriminant and Witness Directions

N. Popescu-Bodorin[*], *Member*, V.E. Balas[**], *Senior Member*, and I.M. Motoc[*], *Student Member*, IEEE
[*]Artificial Intelligence & Computational Logic Lab., Math. & Comp. Sci. Dept., 'Spiru Haret' University, Bucharest, Romania
[**]Faculty of Engineering, 'Aurel Vlaicu' University, Arad, Romania
bodorin@ieee.org, balas@drbalas.ro, motoc@irisbiometrics.org

*Abstract*- **The main topic discussed in this paper is how to use intelligence for biometric decision defuzzification. A neural training model is proposed and tested here as a possible solution for dealing with natural fuzzification that appears between the intra- and inter-class distributions of scores computed during iris recognition tests. It is shown here that the use of proposed neural network support leads to an improvement in the artificial perception of the separation between the intra- and inter-class score distributions by moving them away from each other.**

I. INTRODUCTION

The relation between fuzziness and intelligence is an open problem these days. Fuzzy instruments are usually being used to attempt intelligent problem solving in conditions of incertitude / imprecision and this is also the case discussed here. The main topic of this paper is how to use intelligence in order to achieve biometric decision defuzzification. A neural training model is proposed here as a possible solution for dealing with natural fuzzification that appears between the intra- and the inter-class score distributions computed during iris recognition tests. Are the sets of iris codes somehow separable in a neural perspective? Are the genuine and imposter pairs two separable classes in some space? Is there a neural network structure able to decrease the degree of confusion between inter- and intra-class distributions of scores? It is shown here that using neural-network support leads to an improvement in the artificial perception of the separation between intra- and inter-class distributions of scores by 'moving' the two score distributions away from each other.

Usually in biometric identification / verification, the separation between intra- and inter-class distributions of scores is vague (Fig. 1 in [3]). Even when working on an ideal iris image database [9] this fuzzification is inherent. There are four main categories of factors leading to the fuzzification of the two score distributions: firstly, the acquisition and segmentation conditions, secondly, the feature encoding and feature matching conditions, thirdly, the different posture of the eye relative to the camera, and last but not least, the fact that the laws of radial iris movement are, in fact, unknown, and therefore, successful matching of two samples taken for the same iris is far from being guaranteed when pupil is captured at very different dilations in those two samples.

Each time when the recognition system negotiates between speed and accuracy, if a degree of imprecision is accepted as a counterbalance for gaining speed processing, fuzzification of the two classes of scores is guaranteed. The same situation occurs when the system is not endowed with suitable methods enabling successful recognition of the same iris captured in different acquisition conditions.

The fuzzification between intra- and inter-class scores is usually (and paradoxically) expressed through a crisp concept, namely the Equal Error Rate (EER, [2], [3]). The existence of such a crisp point was not confirmed in our previously undertaken iris recognition tests ([3] - [5]). Indeed, it can be seen in the mentioned references (especially in Fig. 2 and Fig. 3 from [3]) that EER point varies from one recognition test to another and, in fact, the experimental measurement corresponding to the theoretical concept of EER is a fuzzy EER interval - a collection of recognition thresholds for which it is very hard (or simply impossible) to say for sure if they are recognition scores rather than rejection scores or vice-versa. It can be said that in the fuzzy EER interval, the recognition and rejection are fuzzy (vague / imprecise / almost / quasi) equal probable. In the terms proposed and discussed in [10], the fuzzy EER interval (f-EER) is the f-geometry corresponding to the crisp (but theoretical) prototype EER. In terms of logic [6], f-EER corresponds to a third logical state 'u' (unprecisated and uncertain) of the biometric system, different from 0 (which encodes an imposter pair of samples) and 1 (which encodes a genuine pair of samples). It is shown in [5] (see Theorem 2 in [5]) that the logic of such a system is induced by a Boolean algebra of modulo 8 integers. Here in this paper we will further show that despite being unprecisated and uncertain, the fuzzy EER interval (f-EER) is not unprecisable. Defuzzification of f-EER will be achieved here by using an adequate neural network support. In short, the theoretical crisp prototype EER is fuzzified into f-EER by compressing uint8 (8-bit unsigned integer) iris images as binary codes (which are therefore imperfect and incomplete pieces of information, weakened aliases of the original uint8 codes in a space of binary matrices). This operation will be partially reversed by using neural network support in order to recover digital identities as neural memories from the available iris codes.

A. Terminology

In this section we aim to clarify the difference between an *iris code* and a *digital identity*. An iris code is a binary matrix that follows to be recognized (accepted or rejected) as being representative for an *identity* which is a symbolic or numeric data structure associated to a person. In the simplest case, an *identity* is a label - even it is encoded in a numeric vocabulary (like auto-number ID fields). In a little bit more complex scenario, a digital identity is a numeric data structure obtained by detecting and extracting *common* features in a set of samples taken for the same individual, *discriminant* features between the sets of samples taken for different individuals, and by encoding all of these common and discriminant features in a numerical space. Hence a digital identity is a memory that can be trained with iris codes in order to recognize them, or in other words, the digital identity is a recognizer object, whereas the iris code is a recognized object.

A digital identity encodes more entropy than an iris code. As a matrix, it may share the same dimension with the iris code, but if this is the case, then the type of its components will be different (all components having longer binary representation). On the other hand, in a multi-enrollment scenario, the enrolled iris codes together define a digital identity (which therefore contains the same type of components as the iris codes contain, but has bigger dimension, [3]).

In the example that follows to be given in this paper, the digital identities will be double matrices (trained neuronal memories) of the same dimension as the iris codes. A different approach involving neuro-evolutionary trained memories (digital identities) can be found in [5].

### B. Other Neural Approaches to Iris Recognition

The successfully neural approaches to iris recognition are pretty rare indeed. It is somehow explicable because, even this subject is not present at all in scientific publications, AI community in general share a point of view according to which it is very hard to predict where a training procedure deviates from learning features (learning a concept) to learning specific data (memorizing specific instances of a concept). We don't share this view.

On the other hand, why would or should somebody try such a complicate solution for such a simple problem? Actually, the neural networks are simple enough, much simpler than the current state of affairs in the field of iris recognition. For example, it was shown recently [5], [6] that the artificial understanding of iris recognition couldn't be binary and logically consistent simultaneously if the imposter and genuine score distributions collide into each other, whereas a fuzzy 3-valent disambiguated model of iris recognition [6] can guarantee logical consistency. The utility of a neural network is to deconfuse the two score distributions.

## II. HAMMING DISTANCE - A VECTORIAL PERSPECTIVE ON IRIS RECOGNITION

It is not the first time when we say that iris recognition as a field of applied science is still in its childhood. Investigations of iris recognition as a problem of logic [6] and artificial intelligence [4], [5] are indeed very recent topics. On the other hand, in [6] we have shown how important it is the perspective from which iris recognition is viewed and practiced. It is illustrated there (Fig. 1.d in [6]) that improving iris recognition theory and practice depends on searching, identifying and accepting new perspectives over this domain. Another argument sustaining our points of view will be further formulated, proved and explained in this paper: *the present way of using Hamming distance for iris code comparisons is a pure vectorial manner of understanding iris codes*. Hence, similarity of two matrices is decided using vectorial means only. It is not just the fact that representing irides as iris codes is a uint8-to-binary lossy compression but these codes are further compared only as vectors, without taking into account any matrix-type means and properties. In this context could we (or should we) be still surprised about the fact that the imposter and the genuine score distributions (Fig. 1 in [3]) usually collide into each other?

### A. Hamming distance - a pure vectorial manner of understanding and practicing iris code comparisons

Let us consider the simplest case of two unmasked iris codes $IC_1$, $IC_2$ of the same dimension $w \times h$. Their Hamming similarity score (the complement of Hamming distance relative to the unitary score) is:

$$H(IC_1, IC_2) = \frac{sum(IC_1(:) == IC_2(:))}{w \cdot h}. \quad (1)$$

Let us reshape the iris codes $IC_1$, $IC_2$ as the vectors $IC_1(:)$, $IC_2(:)$ of length $\ell = w \cdot h$, and denote $C_{1,2} = (IC_1 == IC_2)$, which is also a vector and will be further referred to as a *comparison code*. The expression of Hamming similarity score becomes:

$$H(IC_1, IC_2) = \frac{sum(IC_1 == IC_2)}{\ell} = \frac{sum(C_{1,2})}{\ell}, \quad (2)$$

and further, if the notation $C_{1,2}$ is simplified to $C$, then:

$$H(C) = \frac{D \cdot C}{D \cdot W} = \frac{\|D\| \cdot C^D}{\|D\| \cdot W^D} = \frac{C^D}{W^D} = \frac{C^D}{\sqrt{\ell}}, \quad (3)$$

where: $D$ and $W$ are further referred to as the *discriminant direction* and the *witness direction*, respectively (in this particular case they are both trivial and equal to the diagonal of the unit hypercube in $\mathbb{R}^\ell$), $C^D$ and $W^D$ are the orthogonal projections of $C$ and $W$ onto $D$, respectively, and '.' signifies the scalar product of two vectors. Hence we've proved the following theorem:

*Theorem 1* (N. Popescu-Bodorin):
*In the binary iris code space $\{0, 1\}^\ell$, the Hamming distance is a purely vectorial feature computable through orthogonal projection of the comparison code onto the main diagonal of the unit hypercube in $\mathbb{R}^\ell$.*

### B. The problem

Now, we are able to see how a crisp (consistent) theory of iris recognition written in terms of discriminant and witness directions would look like. Let us denote as $\mathcal{P}_C$ a conjunction of prerequisite conditions (relative to the image acquisition and processing at all levels from eye image to the iris code) expressed in binary logic, and let $\{D_i\}_{i \in \overline{1,k}}$ a family of nontrivial discriminant directions that need to be established for $k$ enrolled identities, $C = \mathcal{I} \cup \mathcal{G}$ - the partitioning of the comparison code space in imposter comparison codes ($\mathcal{I}$) and genuine comparison codes ($\mathcal{G}$), W - the trivial witness direction, S - a similarity score computed as:

$$S(C) = \frac{C^{D_i}}{W^{D_i}}, \quad (4)$$

for each pair ($IC_i$, $D_i$) which represents the same identity. Regardless the fact that the similarity score could be computed otherwise than in (4), a consistent theory of iris recognition would then say that:

$$\mathcal{P}_C \to [max\{S(C)|C \in \mathcal{I}\} \lneqq min\{S(C)\}|C \in \mathcal{G}], \quad (5)$$

or in other words, if the prerequisite conditions are fulfilled, then the maximum imposter similarity score should be smaller than the minimum genuine similarity score, whereas a comfortable theory of iris recognition would say that:

$$\mathcal{P}_C \to [max\{S(C)|C \in \mathcal{I}\} \ll min\{S(C)|C \in \mathcal{G}\}], \quad (6)$$

or in other words, a safety band can be fitted between the imposter and the genuine similarity score distributions (i.e. the biometric system can be described by the fuzzy 3-valent model of iris recognition, [5], [6]).

The main goal of this paper is showing that discriminant directions and an iris recognition theory of type (5) can be learned heuristically by using neural network support.

## III. PROPOSED METHODOLOGY

Knowing irides and recognizing irides are two very different things. It must be stated clearly if our approach here is intended to announce and describe new iris recognition results or to formulate new points of view about irides and iris codes. In order to improve previously iris recognition approaches or to formulate new iris recognition methodologies, it is necessary to find new knowledge about irides and iris codes. One of our hypotheses is that the confusion between the genuine and imposter score distributions is motivated by the relative position of some iris codes in the iris code space. All iris codes extracted from samples taken for the same eye of the same person are viewed here as clusters in the iris code space, namely *personal clusters*. The separation between the personal clusters is fuzzy, or otherwise the genuine and imposter score distributions should not collide into each other. In this context, defuzzification between the two score distributions should be achievable if each personal cluster would be endowed with a suitable *discriminant charged centroid*, strong enough to alter the space in its immediate proximity by *discriminately attracting* the *members* of the personal cluster and *repulsing* non-members away from the personal cluster. This paper shows that it is possible to learn such special centroids as digital identities, using neural network support.

### A. Iris Segmentation and Encoding

Since neural network training is known to by very sensitive to noise, the segmentation procedure used here is designed to minimize the noise presence in the extracted samples by avoiding the chances that eyelids and eyelashes to escape undetected and unfiltered. On the other hand, neural network training is known to be more expensive when the iris codes grow bigger. For all of these reasons, we choose to work with just a quarter of the actual iris segment, as illustrated in Fig. 1. In this way, iris code dimension and noise presence are both kept to a minimum, whereas signal-to-noise ratio is maximized. Circular Fuzzy Iris Segmentation (CFIS2, [3]) is used in order to extract the iris segment as a circular ring. A quarter of this circular ring (Fig. 1) is unwrapped and further encoded as a binary iris code using Haar-Hilbert encoder [3].

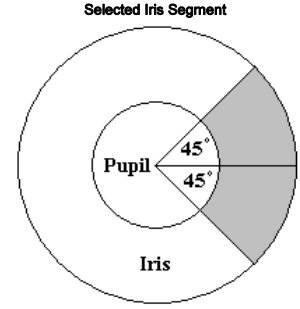

Fig. 1. Iris segmentation: a quarter is selected from each iris sample.

### B. The Learning Algorithm and Neural Network Structure

A well suited neural network for learning discriminant directions is a recurrent neural network which must be compatible with the following space requirements and with the learning algorithm described below. System memory is designed to hold: the current recognition threshold *t*, the safety band *sb*, the numbers of identities *k*, stopping flag *stop* the discriminant directions $\{D_j\}_{j \in \overline{1,k}}$ which are currently trained, the iris codes *IC* on which the current discriminant direction is trained on, and other calibration variables as learning rates *r* and *b*.

**Heuristic Blind Training of Discriminant Directions – HBTDD**
(N. Popescu-Bodorin, V.E. Balas)
1. Initialize *sb*, *t*, *k*, *stop*;
2. Initialize all $\{D_j\}_{j \in \overline{1,k}}$ as random vectors of binary digits
3. Until *stop* do: *stop* = 1 and:
4.    For each comparison code $C_{j,i}$
5.      If $S(C_{j,i})$ is not on the correct side of the safety band:
6.        *stop* = 0;
7.        or $C_{j,i} \in \mathcal{G}$ and then:
8.           $D_j = D_j + r \cdot C_{j,i} - r \cdot \overline{C}_{j,i}$ and *sb* = *sb* - *b*;
9.        or $C_{j,i} \in \mathcal{I}$ and then:
10.         $D_j = D_j - r \cdot C_{j,i} + r \cdot \overline{C}_{j,i}$ and *sb* = *sb* + *b*;
11.    EndIf;
12.    EndFor;
13. EndUntil;
14. END;

In the above heuristic procedure, $\overline{C}_{j,i}$ denotes the binary complement of $C_{j,i}$. HBTDD procedure stops only if all discriminant directions $\{D_j\}_{j \in \overline{1,k}}$ are trained i.e. they produce similarity scores positioned correctly for each comparison code and outside the safety band.

The search is blind in HBTDD because there is no strategy for choosing initial discriminant directions and, most important, the entire procedure ignores the results of any Turing test [8] of iris recognition. Hence, the training of discriminant directions is made by following biometric decisions given by an imprecise software agent (Fig. 1 in [3]), not the accurate biometric decision given by a human agent (Fig. 1.a in [6]). This is why it is considered here that HBTDD works much better than initially expected. Even it is a worst case scenario, an uninformed search in which no specific knowledge about digital identities was used, HBTDD ensures artificial learning of the discriminant directions and artificial understanding of an iris recognition theory of type (5).

### C. Numerical results

The numerical results presented in Fig. 2 and Fig. 3 illustrate an incipient state of training the discriminant directions obtained through HBTDD after 4 iterations, with a safety band of 0.01 in width. It proves that iris code classification based on Hamming similarity can be considered a particular case of iris codes classification using discriminant and witness directions.

It can be seen in Fig. 2 that discriminant directions are *weak digital identities* which diminish the confusion between imposter and genuine score distributions. Still, separation between the two classes of scores is weak illustrating that learning fuzzified prototypes (Fig. 1 in [3]) may reduce the error of classifying imposter and genuine comparison codes but the performances obtained by learning crisp prototypes (Fig. 1.a in [6]) are far much better, [5]. However, as it can be seen in Section V of this paper, the information that the classes of comparison codes are separable in a neural perspective by using discriminant and witness directions can be exploited in the Intelligent Iris Verifier architecture ([4], [5]).

### D. Geometric Interpretation

It can be seen in Fig. 2 that for any enrolled iris code, the ensemble formed by the corresponding discriminant and witness directions act like a lens through which the iris code see its farthest friend (the farthest iris code from the same personal cluster, with which it forms the lowest scored genuine pair) as being closer to him than the nearest enemy (the nearest iris code from any different personal cluster, with which it forms the highest scored imposter pair).

For all enrolled iris codes the difference between the lowest genuine similarity score and the highest imposter similarity score is at least 0.03. Hence, HBTDD is a reliable solution for any personal-use application of iris recognition being able to deliver a kind of nearest-neighbor based biometric decisions which are safer than those based on Hamming distance/similarity.

The recognition function describing a biometric decisional model based on discriminant and witness directions is defined as:

$$\mathcal{R}: \mathcal{D} \times \mathcal{W} \times \mathcal{C} \to [0,1], \quad (7)$$

$$\mathcal{R}(D, W, C) = S(C) = \frac{C^{D_i}}{W^{D_i}} \in [0,1]. \quad (8)$$

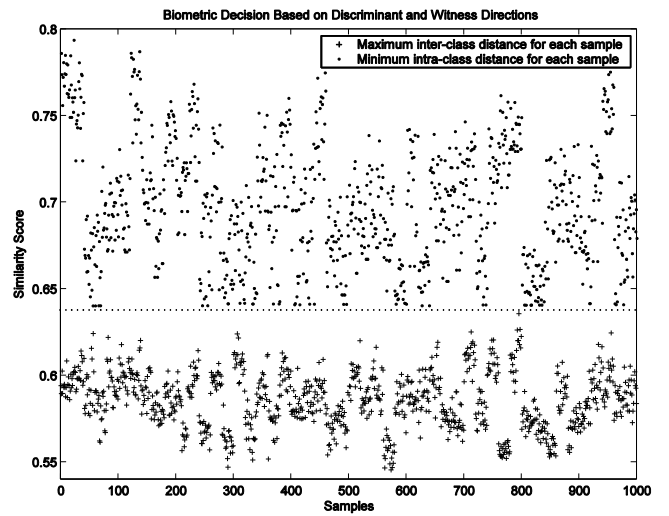

Fig. 2. Statistics of all-to-all comparisons.

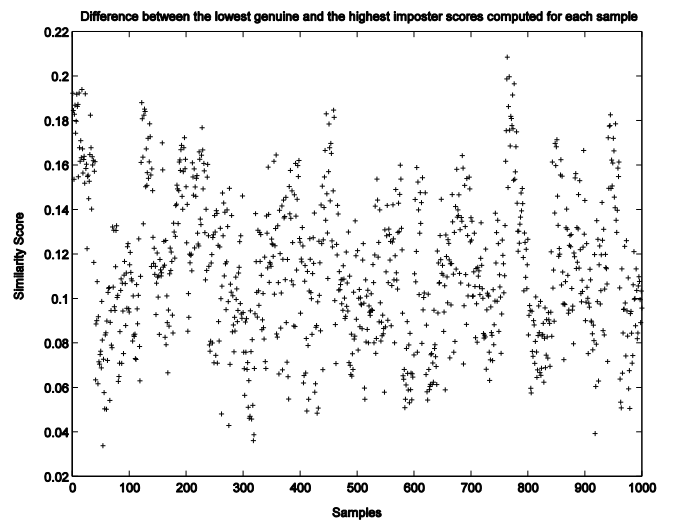

Fig. 3. For each sample, the farthest friend is closer than the nearest enemy.

For each enrolled identity $D$ and for each comparison code let us consider the vector:

$$\bar{\mathcal{R}}(D, W, C) = \mathcal{R}(D, W, C) \cdot \frac{D}{\|D\|} = S(C) \cdot \frac{D}{\|D\|}. \quad (9)$$

If an iris recognition theory of type (5) is learned through discriminant and witness directions, the correspondence:

$$C \mapsto \bar{\mathcal{R}}(D, W, C) \quad (10)$$

maps all imposter comparison codes into a hypersphere within the unit hypersphere in $\mathbb{R}^\ell$ and all genuine comparison codes in between the two hyperspheres.

### IV. IRIS RECOGNITION FORMAL THEORIES

Let us recall that $\mathcal{P}_C$ is a conjunction of prerequisite conditions regarding image acquisition and all image processing steps that must be undertaken in order to generate iris codes, $C$ is a comparison code, $S(C)$ is a similarity score, $\mathcal{I}$ and $\mathcal{G}$ are the sets of imposter and genuine comparison codes respec-tively, and ⊗ denotes logical exclusive disjunction. Above in this paper there are two variants of formal iris recognition theories that have been discussed already, namely:

$$\mathcal{P}_C \to [\max\{S(C)|C \in \mathcal{I}\} \lneqq \min\{S(C)|C \in \mathcal{G}\}], \quad (11)$$

$$\mathcal{P}_C \to [\max\{S(C)|C \in \mathcal{I}\} \ll \min\{S(C)|C \in \mathcal{G}\}]. \quad (12)$$

Let us see now how a formal iris recognition theory should look like if the recognition wouldn't be made by an artificial agent, being made by a human agent instead. We recall that the geometry which illustrates the biometric decisions given by a human agent (Fig. 1.a, in [6]) is a crisp geometry. Hence, the corresponding formal theory of iris recognition is no longer a matter of nuance, degree, incertitude and imprecision regarding the genuine and imposter scores and their distributions, but a crisp theory encodable in binary logic:

$$\mathcal{P}_C \to \{[(C \in \mathcal{I}) \wedge (S(C) \equiv 0)] \otimes [(C \in \mathcal{G}) \wedge (S(C) \equiv 1)]\}, (13)$$

Hence, it makes sense trying to implement software agents enabled to mimic the crisp theory (13), as close as possible, through a fuzzy iris recognition theory:

$$\mathcal{P}_C \to \{[(S(C) \equiv f0) \otimes (S(C) \equiv fu) \otimes (S(C) \equiv f1)]\}, (14)$$

where $f0, fu, f1$ are the fuzzy values of truth within a fuzzy 3-valent disambiguated logical model of iris recognition [6], and:

$$|S^{-1}(fu)| \lll \min(|S^{-1}(f0)|, |S^{-1}(f1)|), \quad (15)$$

i.e. the volume of ambiguous comparison codes is negligible relative to the other two volumes of imposter and genuine comparison codes, respectively.

The list of qualitative conditions imposed while building the theory (14) can be extended with fuzzy rules like 'the safety band must by as wider as possible', or 'the volume of ambiguous comparison codes must be negligible relative to the number of iris codes sampled for a single eye', or regarding the nature and the properties of discriminant directions.

We saw that above, the discriminator directions are *weak digital identities* when their training relies on a fuzzy prototype recognition function and *stationary learning rules*. Their weakness is reflected in the width of the safety band. Hence, it is desirable to train *robust digital identities* able to ensure a wide safety band, a wide gap between the safe imposter and the safe genuine scores. Therefore, our personal list of challenges [4] increases with one:

(C.7.2.) *Find evolutionary methods for encoding enrolled iris codes through robust digital identities determined as trained discriminant directions enabled to give a good approximation for the crisp recognition prototype function previously determined in a Turing test of iris recognition* (N. Popescu-Bodorin).

## V. IIV BALANCED SYSTEM

The results obtained by training discriminant directions (as robust digital identities) on IIV infrastructure [5] are illustrated in Fig. 4 and Fig. 5. The new recognition system obtained in this way is an IIV Balanced System based on learning robust discriminant and witness directions. It is 'balanced' because the values of genuine and imposter absolute safety rates (77.64% vs. 85.19%) are more balanced than the values obtained in the previous IIV simulations [5], [4].

The fuzzy 3-valent disambiguated model [6] corresponding to the IIV Balanced System is shown in Fig. 4.

The eye image database [9] was split into two parts: the training set – containing 5 samples per iris, and the test set which contains 15 samples per iris or less (14) in the cases of failed segmentation (there are 3 cases of failed segmentation between all 1000 images of the database). Then robust discriminant directions have been learned as double matrices on IIV infrastructure from binary iris codes of dimension 64×64 extracted as in Fig. 1.

Fig. 4 and Fig. 5 together illustrate what we call intelligent, consistent and logically argued / motivated biometric safety.

The difference between how different people understand the concepts of '*statistically motivated biometric safety*' and '*logically motivated biometric safety*' is illustrated by a comparison between the results presented in Fig. 4 and Fig. 5 and the results presented in Fig. 1 from [3], or in Fig. 9.a - Fig. 9.f and Table 6 from [2] for the same image database [9], or in Fig. 10 from [1] and in Fig. 4 from [3].

If a biometric system for personal use is detached from IIV Balanced System and endowed with nearest neighbor based biometric decisional support, its safety band would be really wide having 0.4 in width (see Fig. 5).

It can be seen in Fig. 4 that the IIV Balanced System proves to have a crisp understanding of what it means to be a genuine pair (or a genuine comparison code) for 85.19% of all genuine cases (which are scored with crisp unitary recognition score), a crisp understanding of what it means to be an imposter pair (or an imposter comparison code) for 77.64% of all imposter cases (scored with crisp null recognition score), a fuzzy understanding of what it means to be a genuine pair for 14.81% of all genuine cases, (with fuzzy unitary recognition score), a fuzzy understanding of what it means to be an imposter pair for 22.36% of all imposter cases (scored with fuzzy zero recognition score), a global f-consistent ([6], [10]) and complete understanding of iris recognition (being able to give the correct biometric decision for each enrollable pair), a huge safety band (a huge f-EER interval; see the statistics of all-to-all comparisons in Fig. 4) reflecting the artificial consistent understanding of three concepts: 'genuine', 'imposter' and 'unenrollable' (unsafe / uncertain) pair, i.e. the artificial fuzzy 3-valent disambiguated understanding of iris recognition.

The difference between the safety bands obtained through HBTDD and IIV Balanced System is explicable in two ways: firstly, the training tools are more performant in the second case, and secondly, the learned prototype is accurate only in the second case. This illustrates two things: the importance of Turing tests in artificial intelligence (it is the only way of correctly encoding human intelligence in numerical data) and the importance of the informed search. In other words, an informed search (an advanced artificial intelligence tool) targeting a correct prototype will always overcome a blind search targeting an incorrectly chosen prototype.

## VI. CONCLUSION AND FUTURE WORK

In a perfect world, perfect processing tools would exist ensuring that the crisp human perception of iris recognition is perfectly replicable as a crisp artificial perception. In reality, the crisp human perception of iris recognition is expressed as a set of Horn clauses. The world being far from perfect and the processing tools being imprecise enough, the artificial perception of the iris recognition fuzzifies the Horn clauses into fuzzy if-then Sugeno rules [7]. It is a special kind of lossy compression that we would call *semantic compression* and which was previously named, less suggestive, as *fuzzification*. Two actual distinct concepts - namely 'genuine' and 'imposter', represented in a space which is large enough to hold them distinct, are 'mirrored' into a smaller space (are compressed, fuzzified) where their images are no longer distinct. Hence the artificially perceived concepts (mirrored concepts) lost an important part of their original meaning, especially the meaning of their distinct individuality. It is clear now why we said that, in our case, fuzzification means a *lossy compression of meaning*.

This paper showed that, in some conditions, by using tools of artificial intelligence, the memory of distinct individuality can be partially reconstructed / recovered / rediscovered (trained) from a number of compressed samples which taken together as a whole (not individually) host a hidden and apparently lost meaning of the original data.

The reconstruction achieved with IIV Balanced System has so much quality that the recovered memory of distinct individuality allows a wide and comfortable safety band inbetween the numerical representations of the artificial perceived concepts of 'genuine' and 'imposter' comparisons.

Hence, we came up to this point where we showed that 'state of the art' in iris recognition means separating the imposter and genuine score distributions with a wide safety band, not just with the so called recognition threshold. Still, our future works in this field depend almost exclusively on the socio-economical acceptability of this simple and evident truth.

## ACKNOWLEDGMENT

The authors would like to thank Professor Donald Monro (Dept. of Electronic and Electrical Engineering, University of Bath, UK) for granting the access to the Bath University Iris Image Database.

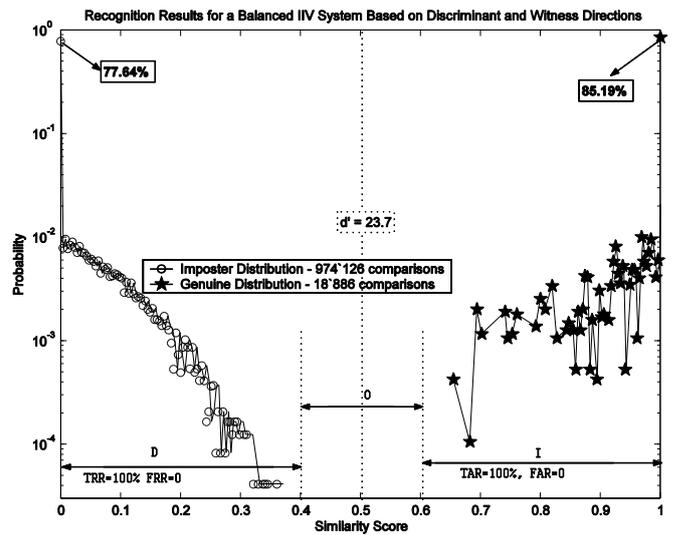

Fig. 4. IIV Balanced System - Statistics of all-to-all comparisons and its fuzzy 3-valent disambiguated logical model.

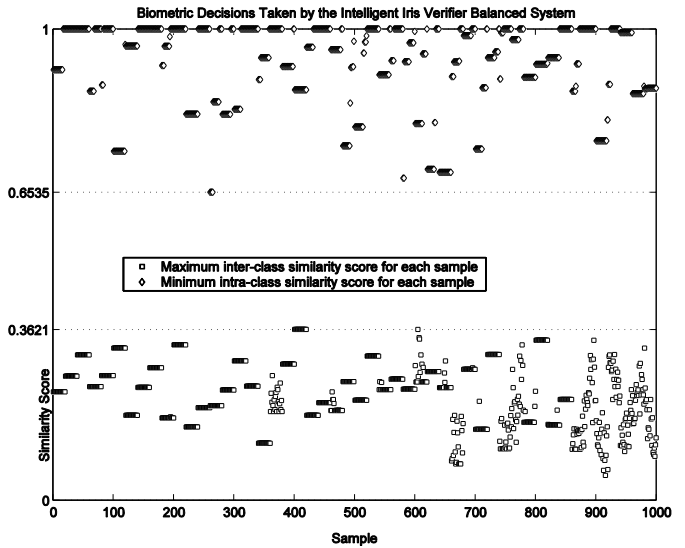

Fig. 5. For each sample of the database [9], the farthest friend is closer than the nearest enemy - the minimum genuine score is much greater than the maximum imposter. The safety band has almost 0.3 in width.